\documentclass[conference,10pt,letterpaper]{IEEEtran}

\IEEEoverridecommandlockouts
\usepackage{cite}
\usepackage{booktabs}
\usepackage{array}
\usepackage{amsmath,amssymb,amsfonts}
\usepackage{algorithmic}
\usepackage{graphicx}
\usepackage{textcomp}
\usepackage{xcolor}
\usepackage{url}
\def\BibTeX{{\rm B\kern-.05em{\sc i\kern-.025em b}\kern-.08em
    T\kern-.1667em\lower.7ex\hbox{E}\kern-.125emX}}
\begin{document}
\bstctlcite{IEEE:explicit-author-names}
\raggedbottom
\clubpenalty=10000
\widowpenalty=10000
\makeatletter
\@clubpenalty=10000 %
\makeatother
\setcounter{dbltopnumber}{1}

\title{Identifying Effective Strategies in Prompt Evolution for Stress and Emotion Prediction}

\author{
\IEEEauthorblockN{SeungHyun Yi\textsuperscript{1,*}, Seungjun Yi\textsuperscript{2,*}, Sriram Sattiraju\textsuperscript{3}, Haeun Park\textsuperscript{4}, Joakim Nguyen\textsuperscript{3}\\
Terence Lim\textsuperscript{5,6}, Joseph Skrovan\textsuperscript{7}, Mehak Beri\textsuperscript{7}, Hitakshi Modi\textsuperscript{8}, Eloy Carrion\textsuperscript{10}\\
Andrew Well\textsuperscript{9}, Carlos M. Mery\textsuperscript{9}, Alexander Rasgon\textsuperscript{11}, Mia K Markey\textsuperscript{2}, Ying Ding\textsuperscript{7,12}}
\IEEEauthorblockA{
\textsuperscript{1}Department of Computer Science, Kyung Hee University, Yongin, South Korea\\
\textsuperscript{2}Department of Biomedical Engineering, \textsuperscript{3}Department of Computer Science, \textsuperscript{4}College of Fine Arts,\\
\textsuperscript{5}College of Natural Sciences, \textsuperscript{7}School of Information, \textsuperscript{8}Department of Pediatrics,\\
\textsuperscript{10}Computer and Data Science, \textsuperscript{11}Psychiatry, \textsuperscript{12}Department of Population Health, Dell Medical School,\\
University of Texas at Austin, Austin, TX, USA\\
\textsuperscript{6}Graphen, Inc., New York, NY, USA\\
\textsuperscript{9}Department of Cardiac Surgery, Division of Pediatric Cardiac Surgery, Vanderbilt University School of Medicine;\\
Pediatric Heart Institute, Monroe Carell Jr. Children's Hospital at Vanderbilt, Nashville, TN, USA}
\thanks{\textsuperscript{*}SeungHyun Yi and Seungjun Yi contributed equally to this work (co-first authors).}
}

\maketitle

\begin{abstract}
Early stress detection and risk prediction are crucial for timely mental health assessment and support. While large language models have recently shown promise for social media stress prediction, relying on accuracy can obscure false positives. Also, prompt revisions yield inconsistent gains across models and tasks, with unclear links to strategy combinations and instruction changes. By comparing six methods on social media stress and emotion data across two model sizes, we found that GEPA and its extension combining instructions from two prompts (GEPA+merge) improved prediction over initial prompts in several settings. We also identified five prompting strategies associated with development gains. These strategies included directing attention to emotional evidence, distinguishing current from past distress, illustrating indirect cues, requiring emotional context, and guiding label selection. Moreover, we documented four observations across tasks and evaluation settings. In a separate set of searches, binary stress prediction improved early, whereas multilabel emotion prediction initially declined and partly recovered. The other three concerned incorrect or conflicting instructions, gains on small comparison samples that did not hold on full development data, and changes in selected prompts' development and reported test rankings. These findings identify effective prompt evolution methods, strategy combinations associated with gains, and task and evaluation limits, providing guidance for applying prompt evolution methods to social media stress detection.

\end{abstract}


\section{Introduction}
\label{sec:introduction}
Early stress detection and risk prediction are crucial for timely mental health assessment and support. Recent studies following adolescents show why stress needs early attention: higher reported stress is associated with later mental health diagnoses in hospital records~\cite{lindholdt2022stress} and depressive feelings~\cite{gniewosz2024stress}. Randomized trials also show that digital stress management and mindfulness programs can reduce employee stress~\cite{heber2016stress,radin2025stress}. In one cohort, observed diagnosis rates over 12 to 16 months were about 4.8\% for adolescents with high stress versus 0.5\% with low stress~\cite{lindholdt2022stress}, while employees receiving guided support averaged about 5 points lower than a waitlist group on a 0 to 40 stress scale after seven weeks~\cite{heber2016stress}.

These needs motivate predicting expressed stress from social media text. Recent work applies large language models (LLMs) to predict stress in Dreaddit Reddit passages using examples~\cite{xu2024mentalllm} or instructions refined with guidance from psychologists~\cite{esmi2025stress}. However, psychologist informed prompt refinement still misclassified about 11\% of passages in one Dreaddit evaluation~\cite{esmi2025stress}. Also, accuracy alone can obscure false positives. Limiting these errors is important because false positives accounted for 90\% of errors during initial prompt evaluation on Dreaddit~\cite{esmi2025stress}. Finally, prompt changes do not consistently improve prediction across models and tasks~\cite{xu2024mentalllm}. These studies leave open how instruction combinations evolve across revisions and relate to gains or declines~\cite{xu2024mentalllm,esmi2025stress}.

To address these limitations, we apply prompt evolution methods for stress and emotion prediction and evaluate accuracy alongside metrics that account for false positives and false negatives. Prompt evolution uses evaluation feedback to revise or combine an LLM's instructions while keeping model weights fixed~\cite{gepa2026}. We trace strategy combinations and recorded instruction revisions alongside development gains and declines, considering how task and evaluation conditions affect their interpretation. These complete prompt comparisons identify associations without isolating individual strategy effects or establishing reductions in false positives or false negatives.

We compared six prompting methods, including prompt evolution, on Dreaddit for stress prediction~\cite{dreaddit2019} and GoEmotions for emotion prediction~\cite{goemotions2020} across two model sizes. We measured overall correctness (accuracy, ACC), performance across labels (Macro-F1), and source support for explanations (Conformability). GEPA achieved the highest mean Macro-F1 on Dreaddit with Qwen3-8B, while GEPA+merge led on GoEmotions with Qwen3-32B. We then performed a detailed analysis of their recorded instruction revisions, supplemented by examples from separate historical searches. This analysis identified five strategies, including focusing on emotional evidence and distinguishing current from past distress, and traced their combinations across revisions. We also documented four observations, including early gains in binary stress prediction but initial declines in multilabel emotion prediction, and gains on small comparison samples that did not hold on full development data. We also discuss how these findings could inform future clinical applications, emphasizing clinician review and validation beyond social media data.

In summary, we ask three questions. \textbf{RQ1:} Are prompt evolution methods effective for stress and emotion prediction tasks? \textbf{RQ2:} Which strategy combinations and instruction changes accompany gains or declines in development performance? \textbf{RQ3:} How do improvements and declines during prompt evolution vary across prediction tasks and evaluation settings?

Our study makes three contributions:
\begin{itemize}
\item \textbf{We identified effective prompt evolution methods for stress and emotion prediction.} Among six evaluated methods, GEPA achieved the highest mean Macro-F1 on Dreaddit with Qwen3-8B (67.81\%), and GEPA+merge on GoEmotions with Qwen3-32B (25.14\%) (Figure~\ref{fig:label-prediction-comparison}).
\item \textbf{We identified five prompting strategies associated with development gains along golden routes for stress and emotion prediction.} These strategies direct attention to emotional evidence, distinguish current from past distress, illustrate indirect cues, require emotional context, and guide emotion label selection (Table~\ref{tab:observed-strategies}). Golden routes, the paths from initial prompts to prompts selected at search end, combine several strategies, with performance measured for complete prompts.
\item \textbf{We documented four observations on task differences, instruction errors, and evaluation limits in prompt evolution for stress and emotion prediction.} Four observations cover these task differences, incorrect or conflicting instructions, gains on small comparison samples that did not hold on full development data, and changes in selected prompts' rankings between development and reported test data.
\end{itemize}

\newcommand{\insertdatasetstatistics}{\begin{table}[!t]
\caption{\textbf{Dataset Statistics}}
\label{tab:dataset-overview}
\centering
\footnotesize
\setlength{\tabcolsep}{3pt}
\renewcommand{\arraystretch}{1.12}
\begin{tabular*}{\columnwidth}{@{\extracolsep{\fill}}lrr@{}}
\toprule
Statistic & Dreaddit~\cite{dreaddit2019} & GoEmotions~\cite{goemotions2020} \\
\midrule
Text unit & Five sentences & One comment \\
Mean words (SD) & 85.6 (32.1) & 12.8 (6.7) \\
Median words & 80 & 12 \\
Annotated items & 3,553 & 58,009 \\
Released classification subset & 3,553 & 54,263 \\
Available labels & 2 & 28 \\
Labels per item & One & One or more \\
\bottomrule
\end{tabular*}
\par\smallskip
\raggedright
Word counts split text on whitespace and cover all released classification items, not the study samples. SD denotes standard deviation. GoEmotions' released classification subset contains items with at least two annotators agreeing on at least one label~\cite{goemotionsdata}.
\end{table}
}
\begin{figure*}[!t]
\centering
\begin{minipage}{\textwidth}
\includegraphics[width=\linewidth]{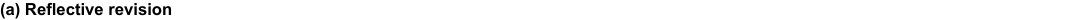}
\end{minipage}\par\smallskip
\includegraphics[width=\textwidth]{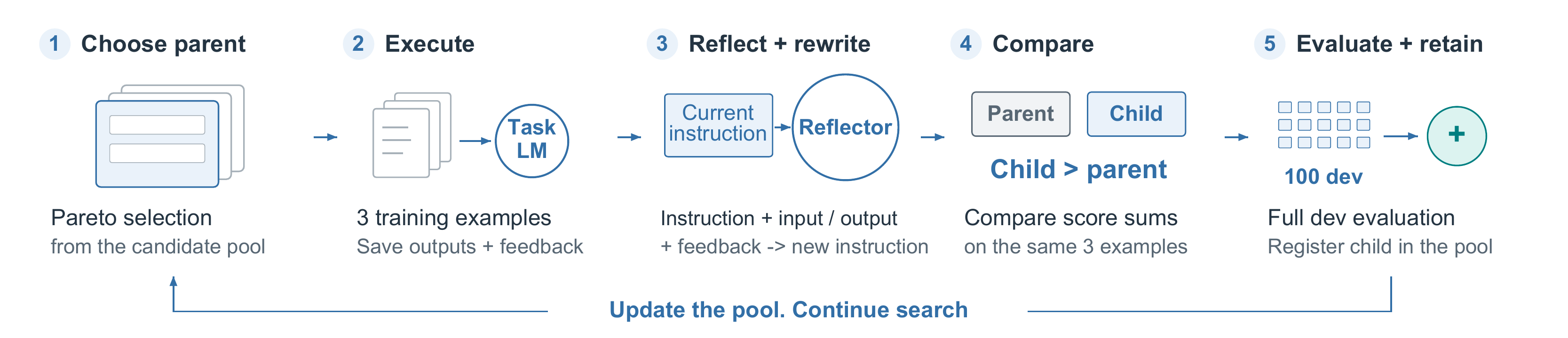}
\par\medskip
\begin{minipage}{\textwidth}
\includegraphics[width=\linewidth]{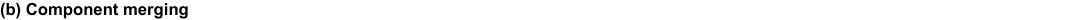}
\end{minipage}\par\smallskip
\includegraphics[width=\textwidth]{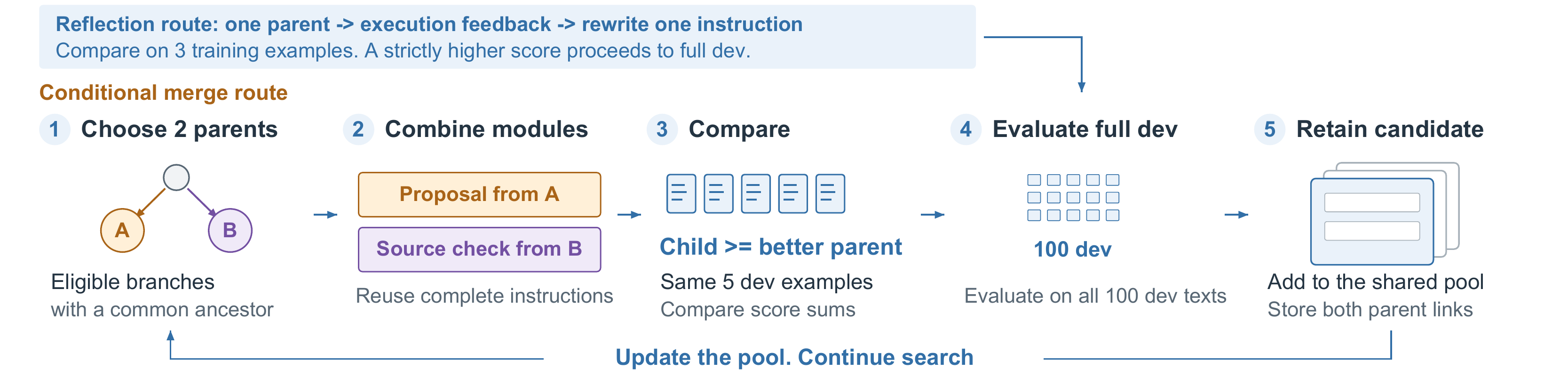}
\caption{\textbf{Prompt evolution through revision and merging.} Main study procedures. (a) GEPA revises one instruction and retains a revision only if its score improves on the same three training examples. (b) GEPA+merge also combines instructions from two parents, accepting a score at least as high as the better parent's on five development examples. Both evaluate accepted prompts on all 100 development examples and select the highest scoring prompt at search end.}
\label{fig:framework}
\label{fig:merge-search}
\end{figure*}

\section{Related Work}
\label{sec:related_work}
\subsection{Stress Detection in Social Media}
Stress detection must distinguish the author's distress from topics, other people, and past experiences. Dreaddit labels Reddit passages~\cite{dreaddit2019}, whereas Guntuku et al. relate users' Facebook and Twitter language to survey measures of perceived stress~\cite{guntuku2019stress}, measuring users rather than individual passages. Turcan et al. incorporate emotion supervision into stress detection, obtaining performance comparable to BERT and using emotion predictions for interpretation~\cite{turcan2021emotion}. Precision and recall distinguish false positives from false negatives, which a single F1 score cannot resolve. Turcan et al. describe BERT missing an author's distress about a friend's situation, while emotion informed models use crying as evidence of stress~\cite{turcan2021emotion}. Xu et al.'s Mental-LLM study reports false positives when FLAN-T5, GPT-3.5, and GPT-4 interpret a calmly recalled anxious job interview as current stress~\cite{xu2024mentalllm}. We compare overall scores across prompt revisions without measuring changes in false positive or false negative rates.

\subsection{Prompt Revision and Health Text Interpretation}
Prompt optimization revises instructions without changing model parameters. TextGrad uses written feedback, MIPROv2 searches instructions and examples, and GEPA uses reflection and component merging~\cite{textgrad2025,mipro2024,dspymiprov2docs,gepa2026}. For clinical adaptation and interpretation, TAMA coordinates generation, evaluation, and revision with clinician oversight~\cite{tama2026}. SFT-TA uses supervised fine tuning to train specialized coding agents, changing model parameters~\cite{sftta2025}. Auto-TA coordinates generation and feedback agents, describing optional reinforcement learning from human feedback as work in progress~\cite{autota2025}. Yi et al.'s clinical framework refines codebooks and records source evidence and analytic decisions~\cite{clinicalta2026}. The clinical position paper emphasizes validity, reliability, and interpretability~\cite{positionta2025}. These principles support tracing evolving instructions and evidence, with clinical validation separate from benchmark prediction.

\subsection{Clinical Context and Patient Simulation}
O'Connor and Kirtley's integrated motivational-volitional model distinguishes the emergence of suicidal ideation from the transition to suicidal behavior~\cite{oconnor2018imv}. Sloan et al.'s randomized noninferiority trial compared five written exposure therapy sessions with 12 cognitive processing therapy sessions in 126 adults with posttraumatic stress disorder (PTSD)~\cite{sloan2018wet}. Craske et al. reviewed exposure therapy through inhibitory learning and retrieval across time and contexts~\cite{craske2008inhibitory}. Hoang et al.'s PSI-Bench preprint assesses depression patient simulators at turn, dialogue, and population levels. They evaluated seven language models across two simulation frameworks and compared benchmark results with mental health professionals' judgments~\cite{hoang2026psibench}.

\newcommand{\insertperformancefigure}{\begin{figure*}[t]
\centering
\includegraphics[width=\textwidth]{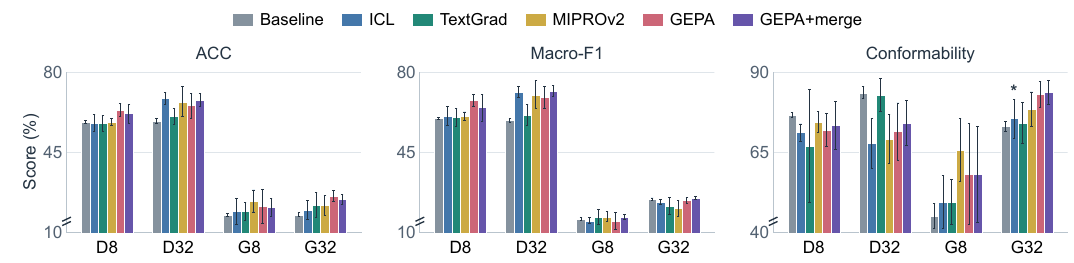}
\caption{\textbf{Label prediction performance across six methods.} Section~\ref{sec:compared-methods} defines all six methods. Groups pair dataset (D: Dreaddit, G: GoEmotions) with Qwen3 model size (8B or 32B). Scores use 200 attempted test items per seed (42 to 46). Bars show means and whiskers one sample standard deviation ($n=5$, except asterisked G32 ICL Conformability, $n=4$). Axes are zoomed, with breaks marking nonzero starts. Compare gaps only within each metric.}
\label{fig:label-prediction-comparison}
\end{figure*}

}
\newcommand{\inserttaskdatasets}{

Table~\ref{tab:dataset-overview} summarizes the two English Reddit datasets. Dreaddit passages come from longer posts, and their binary label (0: no stress, 1: stress) concerns stress expressed by the writer with a negative attitude, rather than a stressful topic~\cite{dreaddit2019}. GoEmotions' neutral label applies when annotators are unsure any emotion is expressed~\cite{goemotions2020}. Neither is a clinical cohort. These benchmarks do not evaluate detection before stress onset, prediction of future disorders, or clinical validity.

Each dataset supplied 100 training and 100 development items, with all splits fixed across methods, models, and executions. Repeated executions measured variability on fixed samples. Prompt evolution analysis used seeds 42 to 44, a subset of final performance executions. Figure~\ref{fig:label-prediction-comparison} specifies the test sample and execution seeds.

Dreaddit had 98 training, 98 development, and 191 test source posts without split overlap. Prepared GoEmotions data lacked author or thread grouping for equivalent checks. The test panel combined 100 historically exposed items with 100 chosen from a fixed ordering independently of outcomes. Results are exploratory and do not evaluate globally untouched data. Records retain split items, seed orders, and source hashes.

}
\section{Methods}
\label{sec:method}
We compared six methods across two model sizes against the initial instruction Baseline using accuracy, Macro-F1, and explanation support (RQ1). Leading results motivated analysis of 24 GEPA and GEPA+merge searches for strategy combinations and instruction changes associated with development gains or declines (RQ2) and differences across tasks and evaluation settings (RQ3), supplemented by six historical searches.

\begin{table}[!t]
\caption{\textbf{Dataset Statistics}}
\label{tab:dataset-overview}
\centering
\footnotesize
\setlength{\tabcolsep}{3pt}
\renewcommand{\arraystretch}{1.12}
\begin{tabular*}{\columnwidth}{@{\extracolsep{\fill}}lrr@{}}
\toprule
Statistic & Dreaddit~\cite{dreaddit2019} & GoEmotions~\cite{goemotions2020} \\
\midrule
Text unit & Five sentences & One comment \\
Mean words (SD) & 85.6 (32.1) & 12.8 (6.7) \\
Median words & 80 & 12 \\
Annotated items & 3,553 & 58,009 \\
Released classification subset & 3,553 & 54,263 \\
Available labels & 2 & 28 \\
Labels per item & One & One or more \\
\bottomrule
\end{tabular*}
\par\smallskip
\raggedright
Word counts split text on whitespace and cover all released classification items, not the study samples. SD denotes standard deviation. GoEmotions' released classification subset contains items with at least two annotators agreeing on at least one label~\cite{goemotionsdata}.
\end{table}

\subsection{Task Setup and Datasets}
\label{sec:task-setup}
\label{sec:datasets}

Table~\ref{tab:dataset-overview} summarizes the two English Reddit datasets. Dreaddit passages come from longer posts, and their binary label (0: no stress, 1: stress) concerns stress expressed by the writer with a negative attitude, rather than a stressful topic~\cite{dreaddit2019}. GoEmotions' neutral label applies when annotators are unsure any emotion is expressed~\cite{goemotions2020}. Neither is a clinical cohort. These benchmarks do not evaluate detection before stress onset, prediction of future disorders, or clinical validity.

Each dataset supplied 100 training and 100 development items, with all splits fixed across methods, models, and executions. Repeated executions measured variability on fixed samples. Prompt evolution analysis used seeds 42 to 44, a subset of final performance executions. Figure~\ref{fig:label-prediction-comparison} specifies the test sample and execution seeds.

Dreaddit had 98 training, 98 development, and 191 test source posts without split overlap. Prepared GoEmotions data lacked author or thread grouping for equivalent checks. The test panel combined 100 historically exposed items with 100 chosen from a fixed ordering independently of outcomes. Results are exploratory and do not evaluate globally untouched data. Records retain split items, seed orders, and source hashes.

Qwen3-8B and Qwen3-32B were the language model backbones. Two reusable instructions ran sequentially: the \emph{answer instruction} produced an initial answer containing labels, evidence, and an explanation from the source, then the \emph{source review instruction} checked for unsupported claims and omissions against the source, corrected them, and returned the final answer.

Both modules shared 4,096 tokens for instructions and training demonstrations, counted by the experiment's tokenizer. Fixed task definitions, output schemas, and the current source passage were excluded. Model context windows and output limits were recorded separately per run.

\subsection{Compared Methods}
\label{sec:compared-methods}
\textbf{Baseline} uses the initial instruction pair without demonstrations or search. \textbf{In context learning (ICL)} keeps initial instructions and weights fixed, adding valid training demonstrations to both modules~\cite{brown2020}. It checks the first eight training candidates in seeded order, retaining examples within the shared context limit whose labels match the reference, without development search. \textbf{TextGrad} revises both instruction texts as variables using textual gradients, written feedback to improve outputs~\cite{textgrad2025}. We configured 20 update slots with three training examples each and full evaluation on all 100 development items at steps 0, 4, 8, 12, 16, and 20. Failures or rejections could reduce accepted updates. We selected the saved instruction state with highest available mean development score, choosing the earliest checkpoint on a tie. \textbf{MIPROv2} (MIPRO version 2) searches instruction and demonstration combinations for both modules~\cite{mipro2024,dspymiprov2docs}. Each module had six instruction alternatives, including the original, and at most two demonstrations bootstrapped from model responses that passed task checks on training inputs. We configured the default and five trial combinations for evaluation on all 100 development items. Selection used the highest available full development score. Failed trials could end early.

\textbf{GEPA} (Genetic-Pareto), our principal prompt evolution method, uses model reflection on task outputs and feedback~\cite{gepa2026}. \textbf{GEPA+merge} adds the merging step in Figure~\ref{fig:framework}. Both retain two modules and use the task backbone for feedback and instruction revision. Only reusable prompt content changes, with model weights, task definitions, and reference data fixed. Different budget units do not imply equal model calls, tokens, or runtime. We used TextGrad 0.1.8, DSPy 3.0.3, and GEPA 0.0.7.

\subsection{Prompt Evolution}
A \emph{root prompt} starts a search with its \emph{root instruction}, the initial task wording. Revising a \emph{parent prompt} produces a \emph{child prompt}. Each main study \emph{saved candidate} pairs answer and source review instructions. Roots, accepted revisions, and merges receive identifiers and can become parents. Rejected revisions receive none. GEPA selects parents using performance across development examples and proposes revisions from their instructions, outputs, and feedback~\cite{gepa2026}. Revision and merge acceptance compare total scores on designated samples. Figure~\ref{fig:framework} specifies comparison, acceptance, and evaluation procedures.

GEPA+merge records both parent links for accepted combinations. Its configured merge limit of five counts only fully evaluated, saved combinations under GEPA 0.0.7, excluding empty or rejected attempts. Eligibility does not guarantee a new combination. Both variants targeted 600 individual task score evaluations, recording actual counts because final evaluation could exceed the target. These main study settings differ from the separate development study's revision and acceptance procedures. Both optimize exact label set agreement, assigning invalid outputs the specified failure outcome. Selection uses mean development scores. The selected \emph{frozen prompt} holds instructions and demonstrations fixed during testing. Acceptance, full development, and test scores remain distinct.

We analyzed 24 searches across both model sizes, both datasets, and seeds 42 to 44. They saved 109 candidates: 24 roots and 85 later candidates (80 retained revisions and five merges). Search logs link instruction pairs and parents to decisions, development scores, and final selections. Identical instructions with different identifiers count separately. Incomplete scores remain missing.

\begin{figure*}[t]
\centering
\includegraphics[width=\textwidth]{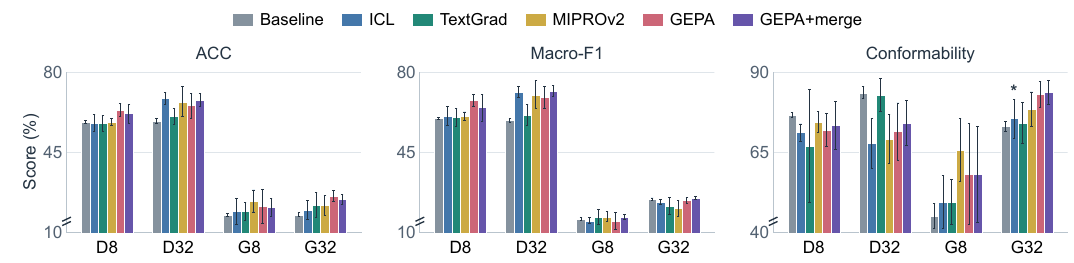}
\caption{\textbf{Label prediction performance across six methods.} Section~\ref{sec:compared-methods} defines all six methods. Groups pair dataset (D: Dreaddit, G: GoEmotions) with Qwen3 model size (8B or 32B). Scores use 200 attempted test items per seed (42 to 46). Bars show means and whiskers one sample standard deviation ($n=5$, except asterisked G32 ICL Conformability, $n=4$). Axes are zoomed, with breaks marking nonzero starts. Compare gaps only within each metric.}
\label{fig:label-prediction-comparison}
\end{figure*}

Search identifiers use dataset/model size/method/seed: Dreaddit/8B/GEPA/44 denotes Dreaddit, Qwen3-8B, GEPA, and random seed 44.

\subsection{Prompt Evolution Analysis}

\label{sec:prompt-analysis}

\subsubsection{Link Search Records}
First, we assembled the analysis records from the 24 main study searches and a separate interactive GEPA record{} containing six additional Qwen3-8B searches for Dreaddit and GoEmotions, seeds 42 to 44, with 135 saved candidates including six roots. In these additional searches, Dreaddit prompts were selected on 100 development items and evaluated unchanged on the reported 100 item test panel. Learned contexts and parent links provide examples, but contexts may omit runtime components, rejected revisions are unavailable, and some development scores are incomplete.

We linked instruction versions to parents, changes, retention decisions, and evaluations to trace inheritance and recurring errors. Yi et al.~\cite{clinicalta2026} similarly make automated thematic analysis traceable through persistent identifiers and recorded operations linking evolving codes and themes to source evidence.

\subsubsection{Code Instruction Content}
Next, a single automated analysis supplied draft codes for distinct module instructions for label prediction, linked to exact passages. No independent human coding or agreement assessment was performed. Seven overlapping functions cover task and decision rules, evidence and grounding, output structure and brevity, examples and cues, verification and correction, scope and granularity, and uncertainty handling. Quotations required for each label combine evidence and output requirements. Task subcategories separate stress decisions from multiple emotion label selection.

From decoded saved instructions and scored label prediction revisions, we counted quotation marks, commas, colons, line breaks, list markers, formatting, and output field notation, excluding fixed task wrappers and source inputs. Retained revisions and corresponding candidates count once, separately from rejected revisions. Source spans identify fields, definitions, decision cues, and quoted labels.

Coding records requested functions, not correctness or evidence of Chain-of-Thought, complete few shot demonstrations, enforced decoding, or calibrated confidence. Negative wording and headings describe expression, not additional functions. Wording changes are distinct from the shared answer generation and review structure and GEPA mutation and merge operations.

\subsubsection{Compare Revisions and Trace Selected Paths}
Exact parent comparisons identified additions, removals, elaborations, replacements, and refinements within unchanged categories. Table~\ref{tab:observed-strategies} links five concrete prompting strategies from the additional searches to datasets, seeds, candidates, instruction changes, gains, and counterexamples. These specific practices differ from the seven broad coding functions. Revision examples follow parent links, while the emotional evidence comparison uses separate branches. The table reports neither frequencies nor golden route counts. We did not calculate new strategy prevalence counts for the 24 main study searches.

\begin{figure*}[!t]
\centering
\begin{minipage}[t]{0.49\textwidth}
\centering
\includegraphics[width=\linewidth]{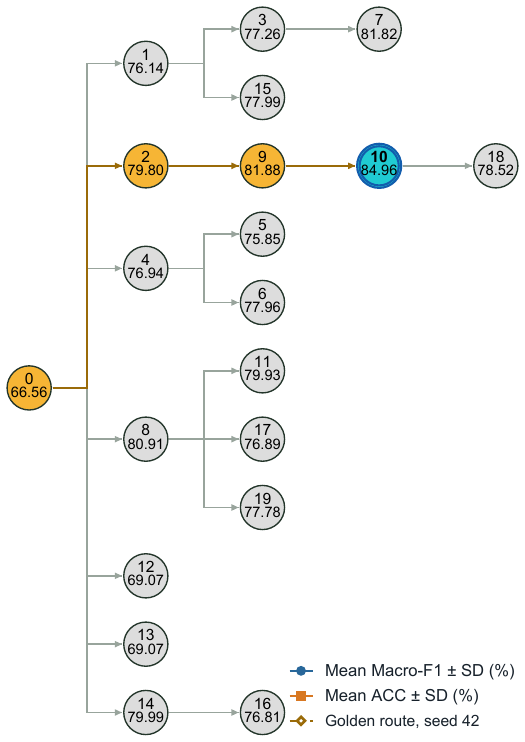}
\par\vspace{2pt}
\includegraphics[width=\linewidth]{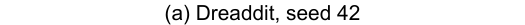}
\par\vspace{2pt}
\includegraphics[width=\linewidth]{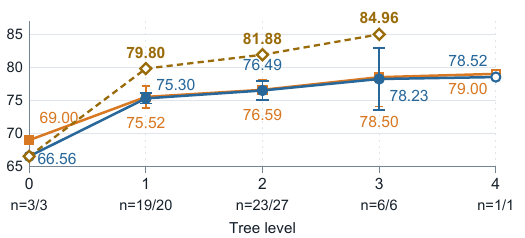}
\end{minipage}\hfill
\begin{minipage}[t]{0.49\textwidth}
\centering
\includegraphics[width=\linewidth]{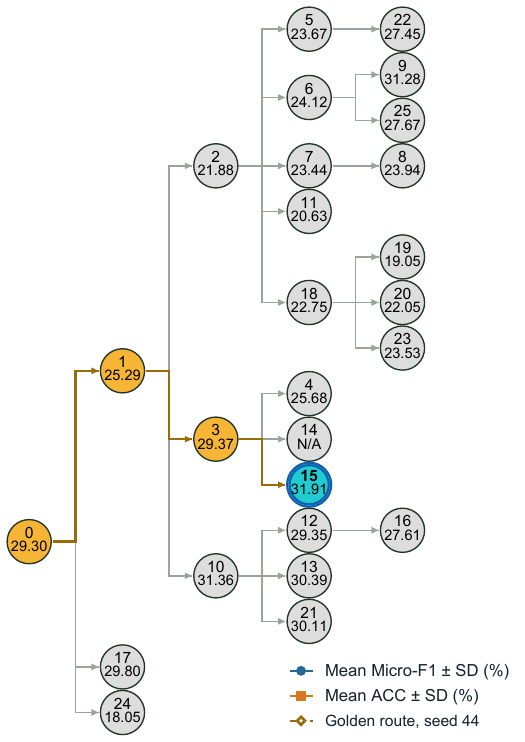}
\par\vspace{2pt}
\includegraphics[width=\linewidth]{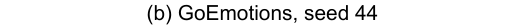}
\par\vspace{2pt}
\includegraphics[width=\linewidth]{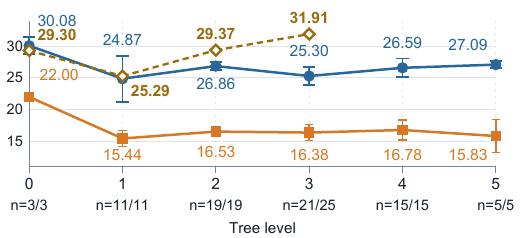}
\end{minipage}
\caption{\textbf{Historical GEPA trees, mean scores, and golden routes.} Qwen3-8B GEPA trees for Dreaddit seed 42 (a) and GoEmotions seed 44 (b) contain 20 and 26 candidates, respectively. Circles show identifiers above development Macro-F1 (a) or Micro-F1 (b), in percent. Arrows link parents to children, amber marks golden routes, cyan circles with double outlines mark selected prompts, and grey circles mark other candidates. N/A denotes missing scores. Positions are adjusted for readability. Blue F1 and orange ACC curves show means across seeds 42 to 44 with one sample SD. At each tree level, $n$ counts scored prompts across seeds in F1/ACC order, while SD is across seed means. Open mean markers indicate one seed without SD. Amber curves show golden route F1 scores without SD. Both plots align vertically with their trees by tree level. Panels use different vertical scales. Candidate Conformability is unavailable. Section~\ref{sec:prompt-analysis} explains the averages and missing scores.}
\label{fig:golden-route}
\end{figure*}

\begin{figure*}[!t]
\centering
\includegraphics[width=\textwidth]{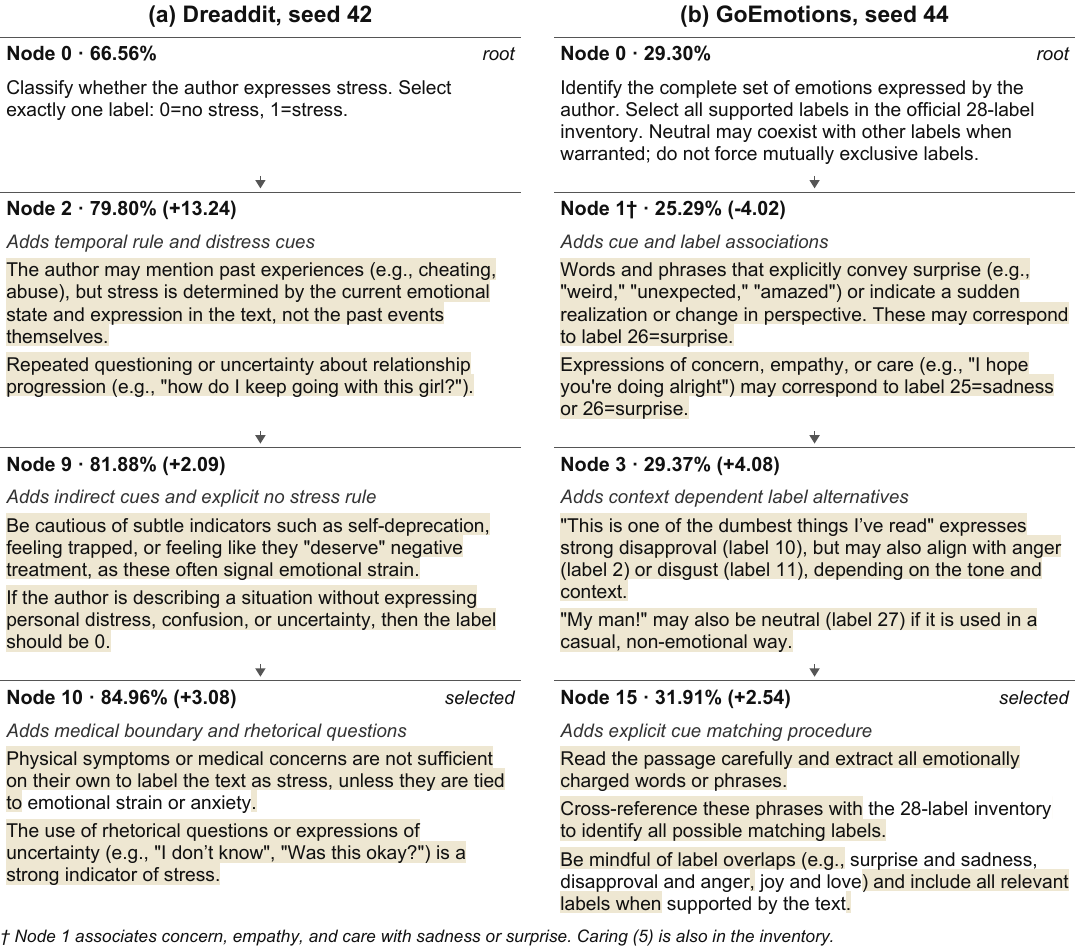}
\caption{\textbf{Prompt changes along historical GEPA golden routes.} Panels match Figure~\ref{fig:golden-route}'s Qwen3-8B trees. Roots are complete and child excerpts retain source wording. Shading marks new or rewritten wording relative to the full parent, not new strategy types. Arrows link parents and children. Scores are development Macro-F1 (a) and Micro-F1 (b), in percent. Parentheses give changes in percentage points from unrounded parent scores. The dagger flags a cue association for scrutiny, not incorrect label identifiers~\cite{goemotionslabelmap}. Section~\ref{sec:prompt-analysis} explains the comparison limits.}
\label{fig:golden-route-excerpts}
\end{figure*}

We traced \emph{golden routes} along parent links, not revision attempt order, from root to selected prompt. Routes guarantee neither stepwise improvement nor the best possible prompt. Figure~\ref{fig:golden-route} shows the additional searches' routes $0 \rightarrow 2 \rightarrow 9 \rightarrow 10$ for Dreaddit (a) and $0 \rightarrow 1 \rightarrow 3 \rightarrow 15$ for GoEmotions (b), with excerpts in Figure~\ref{fig:golden-route-excerpts}. The main study GoEmotions/Qwen3-32B/seed 44 selected merge candidate 3 has two parents, giving paths $0 \rightarrow 1 \rightarrow 3$ and $0 \rightarrow 2 \rightarrow 3$.

\subsubsection{Compare Development Score Patterns}
Finally, we linked complete prompts to full development scores, keeping these distinct from smaller retention comparisons. The additional searches' F1 scores differ from main study exact label agreement. Supplied ACC covers 57 Dreaddit and 78 GoEmotions candidates over 100 development items, versus 52 and 74 with complete F1. Invalid or failed Dreaddit outputs count as errors. GoEmotions ACC comes from the candidate CSV.

\emph{Tree level} counts parent to child steps: root candidate 0 is at level 0, children at 1, and grandchildren at 2. A merge child is one level beyond its deepest parent. For each metric, we average available candidates within each seed and level, then equally weight seed means. Sample standard deviation (SD) uses denominator $s-1$ for $s$ contributing seeds. Missing scores and absent levels are excluded. Only GoEmotions seeds 42 and 43 reach level 5.

Exploratory two sided paired $t$ tests compare F1 across tree levels within each dataset using at least two shared seeds per comparison. Holm correction covers all 21 testable pairs. Tests assume normally distributed differences between independent searches. With at most three seeds and adaptive development selection, tests establish neither causal effects of tree level nor generalization to new data.

\par\addvspace{0.7\baselineskip}\noindent
\begin{minipage}{\columnwidth}
\subsection{Evaluation}
\label{sec:evaluation}
\subsubsection{Label Metrics}
For $N$ texts with reference $Y_i$ and predicted $\widehat Y_i$ label sets, accuracy (ACC) is:
\begin{equation}
\mathrm{ACC}=\frac{100}{N}\sum_{i=1}^{N}\mathbf{1}\{\widehat Y_i=Y_i\}.
\label{eq:accuracy}
\end{equation}
\end{minipage}\par
The indicator is one for an exact binary label match in Dreaddit or complete label set match in GoEmotions. Invalid outputs fail ACC and contribute empty predicted sets to label confusion counts.

For label $\ell$, $TP_\ell$, $FP_\ell$, and $FN_\ell$ count true positives, false positives, and false negatives. We compute
\begin{equation}
F_{1,\ell}=\frac{2TP_\ell}{2TP_\ell+FP_\ell+FN_\ell},\qquad
\mathrm{Macro\text{-}F1}=\frac{100}{L}\sum_{\ell=1}^{L}F_{1,\ell},
\label{eq:macro-f1}
\end{equation}
with $L=2$ for Dreaddit and $L=28$ for GoEmotions. Undefined label F1 is zero. Macro-F1 summarizes final testing, while prompt search optimizes exact label set correctness.

For valid Dreaddit outputs, stress is positive: $TP$ and $TN$ count correct stress and no stress predictions, $FP$ false alarms, and $FN$ missed stress. Precision $TP/(TP+FP)$ is the correct fraction of stress predictions, sensitivity (recall) $TP/(TP+FN)$ the detected fraction of annotated stress, and specificity $TN/(TN+FP)$ the correctly identified fraction of no stress. The false positive rate $FP/(FP+TN)$ is the fraction of annotated no stress passages classified as stress. Macro-F1 weights both classes equally for benchmark comparison but does not encode unequal error costs or establish early detection or clinical utility. Dreaddit's original evaluation also reports precision and recall~\cite{dreaddit2019}.

Dreaddit's original corpus is nearly balanced, with 52.3\% stress passages~\cite{dreaddit2019}. Its published majority baseline predicts stress for every passage, achieving 68.08\% stress class F1~\cite{dreaddit2019} and zero no stress F1. Our two class averaging definition yields approximately 34.04\% Macro-F1. This illustrative calculation uses a published baseline, not our experiments. Macro-F1 thus exposes failure on the unpredicted class despite differing class frequencies.

Invalid outputs produce $FN$ for each reference label and no opposite-label $FP$. They require separate counts alongside valid binary confusion matrices. Main study summaries lack these counts, preventing rankings by missed stress or false alarms.

\subsubsection{Conformability}
\emph{Conformability} is the percentage of attempted test predictions whose explanations pass the source grounding rubric. The common Qwen3-32B judge sees the task, source, and answer, but not the method, task model identity, seed, reference labels, or optimization history. The rubric assesses factual support, evidence relevance, attribution, and absence of material unsupported assertions. Passing requires the maximum score on every dimension. Invalid predictions fail. Unresolved judgments are not dropped from the denominator and leave the run's Conformability undefined. The measure differs from label accuracy and human validation of an interpretation.

\subsubsection{Reporting Protocol}
We evaluated 120 conditions, each contributing one evaluated attempt. GoEmotions/Qwen3-32B/ICL seed 44 had an unresolved Conformability judgment but contributed ACC and Macro-F1 scores. Execution records retain runtime revisions, earlier failures, and undefined metrics. Differences in output limits and failure handling prevent isolating method or prompting strategy effects.

\section{Results and Discussion}
\label{sec:results}
\label{sec:discussion}
\label{sec:search-analysis}

\subsection{RQ1: Are Prompt Evolution Methods Effective for Stress and Emotion Prediction Tasks?}
\label{sec:rq1}

GEPA and GEPA+merge improved mean stress prediction accuracy and Macro-F1 over Baseline on Dreaddit. With Qwen3-8B, GEPA achieved the highest Macro-F1 ($67.81 \pm 2.58$) and ACC ($63.50 \pm 2.78$), but lower Conformability than Baseline (Figure~\ref{fig:label-prediction-comparison}). With Qwen3-32B, GEPA+merge led Macro-F1 (71.80\%) and ICL led ACC (68.50\%). On GoEmotions, GEPA+merge exceeded Baseline in mean accuracy, Macro-F1, and Conformability with both model sizes, although leaders varied by model and metric. With Qwen3-32B, GEPA+merge led Macro-F1 ($25.14 \pm 0.61$) and Conformability ($83.80 \pm 3.70$). GEPA led exact label set ACC ($26.00 \pm 2.42$), but its Macro-F1 remained below Baseline's. With Qwen3-8B, MIPROv2 led Macro-F1 (16.82\%) and ACC (23.70\%).

We next examine GEPA and GEPA+merge strategy combinations and instruction changes accompanying development gains or declines (RQ2), and how these vary across prediction tasks and evaluation settings (RQ3).

\begin{table*}[!t]
\caption{\textbf{Effective Strategies in Prompt Evolution: Supporting Examples}}
\label{tab:observed-strategies}
\label{tab:prompting-strategies}
\centering
\small
\setlength{\tabcolsep}{4pt}
\renewcommand{\arraystretch}{1.05}
\setlength{\aboverulesep}{1.5pt}
\setlength{\belowrulesep}{1.5pt}
\begin{tabular*}{\textwidth}{@{\extracolsep{\fill}}>{\raggedright\arraybackslash}p{0.23\textwidth}>{\raggedright\arraybackslash}p{0.465\textwidth}>{\raggedright\arraybackslash}p{0.265\textwidth}@{}}
\toprule
\textbf{Strategy} & \textbf{Examples from prompts} & \textbf{Development scores (\%)} \\
\midrule
\textbf{Directing Attention to Emotional Evidence}\par\vspace{1pt}
\textit{Dreaddit, seed 42} & \textit{Candidate 1:} ``the label is based on the author's emotional state, not on the topic''\par\vspace{1pt}
\textit{Candidate 10:} ``Stress is primarily linked to the emotional state of the author, not the external situation itself.'' & Candidate 1: 76.14\par Candidate 10: 84.96\par\vspace{1pt} Separate branches\par Candidate 10 selected \\
\midrule
\textbf{Distinguish current from past distress} \textit{Dreaddit, seed 42} & ``stress is determined by the current emotional state and expression in the text, not the past events themselves.'' & Candidates $0 \rightarrow 2$\par $66.56 \rightarrow 79.80$ \\
\midrule
\textbf{Illustrate indirect emotional cues} \textit{Dreaddit, seed 42} & ``self-deprecation'', ``feeling trapped'', and ``feeling like they `deserve' negative treatment'' & Candidates $2 \rightarrow 9$\par $79.80 \rightarrow 81.88$ \\
\midrule
\textbf{Require emotional context}\par\vspace{1pt}
\textit{Dreaddit, seed 42} & ``Physical symptoms or medical concerns are not sufficient on their own to label the text as stress, unless they are tied to emotional strain or anxiety.'' & Candidates $9 \rightarrow 10$\par $81.88 \rightarrow 84.96$\par\vspace{1pt} Child 18: 78.52 (Same rule retained) \\
\midrule
\textbf{Guide emotion label selection}\par\vspace{1pt}
\textit{GoEmotions} & \textit{Neutral:} ``should only be selected if the text explicitly reflects no strong emotional content.'' 1. ``extract all emotionally charged words or phrases.'' 2. ``Cross-reference these phrases with the 28-label inventory to identify all possible matching labels.'' & Seed 43, candidates $0 \rightarrow 3$\par $31.62 \rightarrow 24.18$\par\vspace{2pt} Seed 44, candidates $3 \rightarrow 15$\par $29.37 \rightarrow 31.91$ \\
\bottomrule
\end{tabular*}
\par\smallskip
\raggedright\footnotesize
Verbatim excerpts from the separate historical Qwen3-8B searches. Scores are development Macro-F1 for Dreaddit and Micro-F1 for GoEmotions. Arrows link parents and children. The first strategy compares separate branches. Scores evaluate whole prompts, not individual rules.
\end{table*}

\subsection{RQ2: Which Strategy Combinations and Instruction Changes Accompany Gains or Declines in Development Performance?}
\label{sec:rq2}

\label{sec:strategy-content}

Table~\ref{tab:observed-strategies} illustrates five strategies with saved prompt excerpts and development scores from golden routes and other branches. The examples show changes and exceptions, not frequency rankings. Dreaddit rules distinguish the author's emotion from topics or others' experiences, current distress from past or resolved events, and distress from affection, help seeking, or coping.

\textbf{Strategy 1) Directing attention to emotional evidence} prioritizes the writer's expressed state over difficult circumstances alone.

\textcolor{black}{All three selected historical Qwen3-8B Dreaddit prompts distinguish the writer's emotion from the topic. Seed 42 candidate 1 requires that ``the label is based on the author's emotional state, not on the topic''. Its development Macro-F1 is 76.14\%, versus 84.96\% for selected candidate 10 on a separate branch (Table~\ref{tab:observed-strategies}). Seed 43 candidate 11 instructs: ``Focus on emotional tone and language, not the content of the passage itself''. Seed 44 candidate 10 notes that ``The presence of challenging life events (e.g., family conflict, financial strain) does not automatically imply stress.''} The shared rule alone does not explain the score difference.

In the main Dreaddit/8B/GEPA/44 search, candidate 4 revises candidate 1, retaining its strategy categories but expanding emotional cues to behavioral changes and medical context. Development accuracy (ACC) on all 100 items rises from 56\% to 61\%. Emotional evidence occurs in stronger and weaker prompts, emphasizing complete instruction combinations.

\textbf{Strategy 2) Distinguishing current from past distress} clarifies which time period matters. Dreaddit seed 42 candidate 2 adds a current emotional state rule alongside stress criteria and examples (Table~\ref{tab:observed-strategies}). Observation~1 relates this revision to the pattern across tree levels.

\textbf{Strategy 3) Illustrating indirect emotional cues} provides examples of less explicit distress. Dreaddit seed 42 candidate 9 adds putting oneself down, feeling trapped, and deserving negative treatment to candidate 2's cues. Macro-F1 rises from 79.80\% to 81.88\%. These illustrations are shorter than complete demonstrations containing a source text and labeled answer.

\textbf{Strategy 4) Requiring emotional context} distinguishes a health concern from distress about it. Historical Dreaddit seed 42 candidate 10 requires emotional strain or anxiety for physical symptoms or medical concerns (Figures~\ref{fig:golden-route} and~\ref{fig:golden-route-excerpts}). Candidate 9 to selected candidate 10 adds physical and emotional decision boundaries, indirect distress examples, and temporal clarification among other edits, while development Macro-F1 rises from 81.88\% to 84.96\%. Child 18 retains this rule but removes five illustrative phrases, tightens a temporal rule, and reorders instructions. Macro-F1 falls to 78.52\%, so retaining the rule is insufficient to preserve the higher score.

\textbf{Strategy 5) Guiding emotion label selection} specifies when neutral or multiple labels apply. Historical GoEmotions seed 44 candidate 15 asks the model to ``extract all emotionally charged words or phrases'', then match them to the inventory and use neutral if no clear emotional tone is present. Its parent already permits overlapping emotions. Table~\ref{tab:observed-strategies} contrasts this procedure with seed 43 candidate 3's restrictive neutral rule.

In GoEmotions/32B/GEPA+merge/44, selected candidate 3 combines candidate 1's answer instruction with candidate 2's source review instruction, both unchanged, creating a new prompt combination (Figure~\ref{fig:framework}(b)).

All main study roots requested grounding, concise output, and answer review, so these were not discoveries. Historical Dreaddit seed 42 candidates 1 and 12 respectively request quotations and explanations, and a label alone. Both revise root 0, not each other. 

Evolved prompts commonly used lists, emphasis, conditional rules, and restrictions. All contained commas. Median combined instruction lengths were 2,842 characters for GEPA and 3,270 for GEPA+merge. Quotation marks and field notation served examples, output definitions, and other purposes without necessarily identifying source evidence. These structural observations do not establish that formatting caused performance differences.

\subsection{RQ3: How Do Improvements and Declines During Prompt Evolution Vary Across Prediction Tasks and Evaluation Settings?}
\label{sec:rq3}
Label number, coexistence, and existing instructions may explain task differences.

\begingroup
\clubpenalties 4 10000 10000 10000 0

\textbf{Observation 1. Score trajectories differ by prediction task.} In the separate search record, mean development F1 across seeds 42 to 44 rose steadily across tree levels for binary stress prediction but initially fell and partly recovered for emotion prediction with 28 labels. Golden routes were similar, with GoEmotions eventually exceeding its root (Figure~\ref{fig:golden-route}). Dreaddit's mean gain from tree level 0 to 1 was 8.73 percentage points across three seeds ($p=0.0026$ unadjusted, $p=0.055$ after 21 comparisons). No adjusted comparison met the 0.05 threshold.\par
\endgroup

\textcolor{black}{In three historical Qwen3-8B searches per dataset, all 19 scored Dreaddit root children gained 1.60 to 14.34 development Macro-F1 percentage points, expanding: ``Classify whether the author expresses stress. Select exactly one label: 0=no stress, 1=stress.'' Seed 42 candidate 2 on the golden route adds that ``stress is determined by the current emotional state and expression in the text, not the past events themselves.'' Macro-F1 rises from 66.56\% to 79.80\%. Beyond tree level 1, ten of 30 scored revisions improved, 19 declined, and one tied. All six scored children of seed 44 candidate 3 fell by 1.54 to 3.64 points. One root child and four later children lacked complete scores, including candidate 3's seventh child.}

\textcolor{black}{GoEmotions' early revisions generally reduced performance: only two of eleven scored root children improved development Micro-F1, each by less than one point, while seven fell by more than four. All five scored seed 43 root children declined. Golden route candidate 3 specifies that neutral ``should only be selected if the text explicitly reflects no strong emotional content,'' while Micro-F1 falls from 31.62\% to 24.18\%.}

\begingroup
\clubpenalty=10000
\textbf{Observation 2. Some automatically revised prompts contained incorrect or conflicting task instructions.} The first retained GoEmotions/8B/GEPA/42 candidate requests a single primary emotion, whereas the task allows multiple supported emotion labels.

In historical Dreaddit seed 42, selected candidate 10 contains overlapping positive and negative conditions despite 84.96\% development Macro-F1. Candidate 15 retains its parent's no stress fallback but adds a competing rule against defaulting to neutral under ambiguity. Dreaddit seed 43 candidate 16 references numbered example feedback without the complete example in its learned context. GoEmotions seed 42 candidate 7 maps joy and gratitude to 15, although their official identifiers are 17 and 15, respectively~\cite{goemotionslabelmap}.

\textcolor{black}{All three historical Qwen3-8B GoEmotions searches include a direct root revision with incorrect label identifiers. Seed 43 candidate 19 has only 3 correct identifiers among 28 entries and adds labels absent from the dataset, including ``stress'' and ``terror.'' Candidate 24 in seed 44 lists ``anger'' under seven identifiers and numbers the inventory from 1 to 28. These are their searches' largest development Micro-F1 declines among scored parent and child comparisons, from 31.62\% to 17.84\% and from 29.30\% to 18.05\%. Seed 42 candidate 6 lists each of 15 names twice under identifiers 0 to 29, yet falls only from 29.30\% to 27.41\%. Retained candidate 6's three descendants (19, 21, and 24) keep this inventory and score 27.54\% to 28.87\%.}\par
\endgroup

\textbf{Observation 3. Comparison gains did not guarantee gains on the full development set.} In GoEmotions/32B/GEPA/44, candidate 3 is retained after improving over parent 2 from zero to one correct item on the three item training comparison. Its full development score is 19\%, versus its parent's 20\%. The final frozen candidate 5 scores 25\% on development data.

\textcolor{black}{The search's first attempted revision discourages multiple labels with overlapping meanings or when one suffices. Parent and revision both score zero on three comparison items. It is rejected without an identifier or full development evaluation.}

\begingroup
\clubpenalties 3 10000 10000 0
\textbf{Observation 4. Among three selected prompts, the best development prompt ranked worst on the reported test.} \textcolor{black}{Among three selected historical Qwen3-8B Dreaddit prompts, seed 42 candidate 10 ranks first on development Macro-F1 (84.96\%) but last on reported test Macro-F1 (75.99\%). Seed 44 candidate 10 ranks last on development Macro-F1 (78.95\%) and first on reported test Macro-F1 (79.99\%).}\par
\endgroup

\section{Clinical Implications and Future Work}
\label{sec:clinical-future-work}

From a clinical perspective, the strategies associated with gains mirror clinical reasoning: distinguishing a person’s expressed distress from difficult circumstances, separating current from past distress, and attending to indirect cues such as feeling trapped. Entrapment is central to suicide theory; the integrated motivational–volitional model proposes that defeat and entrapment drive the emergence of suicidal ideation~\cite{oconnor2018imv}. Because the evolved prompts are written in plain language, clinicians can check them for errors. This checking is essential, since some prompts contained wrong labels and contradictory rules. These methods are not ready for clinical use, because they were tested only on social media posts and the study did not measure how often stress was falsely flagged or missed.

Future work should test these methods in settings where text is already part of care. In written exposure therapy, which was noninferior to cognitive processing therapy for PTSD despite fewer sessions~\cite{sloan2018wet}, patients write repeatedly about the same trauma. Emotion detection could help therapists track change across these narratives. Change across sessions is a better target than intensity within one, because exposure research indicates that neither the amount of fear reduction within a session nor the fear level at its end predicts treatment outcome~\cite{craske2008inhibitory}. Prompts would also need adapting, since a rule that discounts past events conflicts with writing about past trauma. A second direction is simulated patients for clinical training, where emotion detection could check realism; current depression simulators progress through therapeutic stages and toward positive emotion too quickly~\cite{hoang2026psibench}. Finally, text-based stress and emotion detection may strengthen multimodal systems that also use speech and physiological signals, but transfer from social media posts to clinical speech must be tested first.

\section{Conclusion}
\label{sec:conclusion}
Among six evaluated methods, GEPA achieved the highest mean Macro-F1 for Dreaddit stress prediction with Qwen3-8B, and GEPA+merge for GoEmotions emotion prediction with Qwen3-32B. Across both model sizes, accuracy gains did not always coincide with better Macro-F1 or explanation support. Selected paths combined several of the five identified strategies also found in weaker prompts. Apparently useful revisions could reduce performance. Complete instruction combinations need evaluation in context, while isolated strategy effects remain unmeasured.

Early gains in binary stress prediction contrasted with initial declines and partial recovery in multilabel emotion prediction, possibly reflecting task structure and initial instructions. Incorrect or conflicting instructions, comparison gains that failed on full development data, and development versus reported test rank changes further qualified progress. The proposed clinical applications require controlled tests of individual strategies and combinations, measurement of false positives and missed stress, and validation on representative clinical data.

\bibliographystyle{IEEEtran}
\bibliography{references}

\begin{thebibliography}{10}
\providecommand{\url}[1]{#1}
\csname url@samestyle\endcsname
\providecommand{\newblock}{\relax}
\providecommand{\bibinfo}[2]{#2}
\providecommand{\BIBentrySTDinterwordspacing}{\spaceskip=0pt\relax}
\providecommand{\BIBentryALTinterwordstretchfactor}{4}
\providecommand{\BIBentryALTinterwordspacing}{\spaceskip=\fontdimen2\font plus
\BIBentryALTinterwordstretchfactor\fontdimen3\font minus \fontdimen4\font\relax}
\providecommand{\BIBforeignlanguage}[2]{{%
\expandafter\ifx\csname l@#1\endcsname\relax
\typeout{** WARNING: IEEEtran.bst: No hyphenation pattern has been}%
\typeout{** loaded for the language `#1'. Using the pattern for}%
\typeout{** the default language instead.}%
\else
\language=\csname l@#1\endcsname
\fi
#2}}
\providecommand{\BIBdecl}{\relax}
\BIBdecl

\bibitem{lindholdt2022stress}
\BIBentryALTinterwordspacing
L.~Lindholdt, M.~Labriola, J.~H. Andersen, M.-M.~Z. Kjeldsen, C.~Obel, and T.~Lund, ``Perceived stress among adolescents as a marker for future mental disorders: A prospective cohort study,'' \emph{Scandinavian Journal of Public Health}, vol.~50, no.~3, pp. 412--417, 2022. [Online]. Available: \url{https://doi.org/10.1177/1403494821993719}
\BIBentrySTDinterwordspacing

\bibitem{gniewosz2024stress}
\BIBentryALTinterwordspacing
G.~Gniewosz, ``Clarifying the links between perceived stress and depressiveness: a longitudinal study of {COVID-19}'s effects on adolescents in {Germany},'' \emph{Journal of Youth and Adolescence}, vol.~53, no.~10, pp. 2348--2362, 2024. [Online]. Available: \url{https://doi.org/10.1007/s10964-024-02012-8}
\BIBentrySTDinterwordspacing

\bibitem{heber2016stress}
\BIBentryALTinterwordspacing
E.~Heber, D.~Lehr, D.~D. Ebert, M.~Berking, and H.~Riper, ``Web-based and mobile stress management intervention for employees: A randomized controlled trial,'' \emph{Journal of Medical Internet Research}, vol.~18, no.~1, p. e21, 2016. [Online]. Available: \url{https://www.jmir.org/2016/1/e21/}
\BIBentrySTDinterwordspacing

\bibitem{radin2025stress}
\BIBentryALTinterwordspacing
R.~M. Radin, J.~Vacarro, E.~Fromer, S.~E. Ahmadi, J.~Y. Guan, S.~M. Fisher, S.~D. Pressman, J.~F. Hunter, K.~Sweeny, A.~J. Tomiyama, L.~T. Hofschneider, M.~J. Zawadzki, L.~Gavrilova, E.~S. Epel, and A.~A. Prather, ``Digital meditation to target employee stress: A randomized clinical trial,'' \emph{JAMA Network Open}, vol.~8, no.~1, p. e2454435, 2025. [Online]. Available: \url{https://doi.org/10.1001/jamanetworkopen.2024.54435}
\BIBentrySTDinterwordspacing

\bibitem{xu2024mentalllm}
\BIBentryALTinterwordspacing
X.~Xu, B.~Yao, Y.~Dong, S.~Gabriel, H.~Yu, J.~Hendler, M.~Ghassemi, A.~K. Dey, and D.~Wang, ``{Mental-LLM}: Leveraging large language models for mental health prediction via online text data,'' \emph{Proceedings of the ACM on Interactive, Mobile, Wearable and Ubiquitous Technologies}, vol.~8, no.~1, pp. 1--32, 2024. [Online]. Available: \url{https://doi.org/10.1145/3643540}
\BIBentrySTDinterwordspacing

\bibitem{esmi2025stress}
\BIBentryALTinterwordspacing
N.~Esmi, A.~Shahbahrami, Y.~Nabati, B.~Rezaei, G.~Gaydadjiev, and P.~de~Jonge, ``Stress detection through prompt engineering with a general-purpose {LLM},'' \emph{Acta Psychologica}, vol. 260, p. 105462, 2025. [Online]. Available: \url{https://doi.org/10.1016/j.actpsy.2025.105462}
\BIBentrySTDinterwordspacing

\bibitem{gepa2026}
\BIBentryALTinterwordspacing
L.~A. Agrawal, S.~Tan, D.~Soylu, N.~Ziems, R.~Khare, K.~Opsahl-Ong, A.~Singhvi, H.~Shandilya, M.~J. Ryan, M.~Jiang, C.~Potts, K.~Sen, A.~Dimakis, I.~Stoica, D.~Klein, M.~Zaharia, and O.~Khattab, ``{GEPA}: Reflective prompt evolution can outperform reinforcement learning,'' in \emph{International Conference on Learning Representations}, 2026. [Online]. Available: \url{https://proceedings.iclr.cc/paper_files/paper/2026/hash/0e9e708b6f48e14fd0ac29e167413f76-Abstract-Conference.html}
\BIBentrySTDinterwordspacing

\bibitem{dreaddit2019}
\BIBentryALTinterwordspacing
E.~Turcan and K.~McKeown, ``{Dreaddit}: A {Reddit} dataset for stress analysis in social media,'' in \emph{Proceedings of the Tenth International Workshop on Health Text Mining and Information Analysis ({LOUHI} 2019)}.\hskip 1em plus 0.5em minus 0.4em\relax Association for Computational Linguistics, 2019, pp. 97--107. [Online]. Available: \url{https://aclanthology.org/D19-6213/}
\BIBentrySTDinterwordspacing

\bibitem{goemotions2020}
\BIBentryALTinterwordspacing
D.~Demszky, D.~Movshovitz-Attias, J.~Ko, A.~Cowen, G.~Nemade, and S.~Ravi, ``{GoEmotions}: A dataset of fine-grained emotions,'' in \emph{Proceedings of the 58th Annual Meeting of the Association for Computational Linguistics}.\hskip 1em plus 0.5em minus 0.4em\relax Association for Computational Linguistics, 2020, pp. 4040--4054. [Online]. Available: \url{https://aclanthology.org/2020.acl-main.372/}
\BIBentrySTDinterwordspacing

\bibitem{guntuku2019stress}
\BIBentryALTinterwordspacing
S.~C. Guntuku, A.~Buffone, K.~Jaidka, J.~C. Eichstaedt, and L.~H. Ungar, ``Understanding and measuring psychological stress using social media,'' \emph{Proceedings of the International AAAI Conference on Web and Social Media}, vol.~13, no.~1, pp. 214--225, 2019. [Online]. Available: \url{https://ojs.aaai.org/index.php/ICWSM/article/view/3223}
\BIBentrySTDinterwordspacing

\bibitem{turcan2021emotion}
\BIBentryALTinterwordspacing
E.~Turcan, S.~Muresan, and K.~McKeown, ``Emotion-infused models for explainable psychological stress detection,'' in \emph{Proceedings of the 2021 Conference of the North American Chapter of the Association for Computational Linguistics: Human Language Technologies}.\hskip 1em plus 0.5em minus 0.4em\relax Association for Computational Linguistics, 2021, pp. 2895--2909. [Online]. Available: \url{https://aclanthology.org/2021.naacl-main.230/}
\BIBentrySTDinterwordspacing

\bibitem{textgrad2025}
\BIBentryALTinterwordspacing
M.~Yuksekgonul, F.~Bianchi, J.~Boen, S.~Liu, P.~Lu, Z.~Huang, C.~Guestrin, and J.~Zou, ``Optimizing generative {AI} by backpropagating language model feedback,'' \emph{Nature}, vol. 639, no. 8055, pp. 609--616, 2025. [Online]. Available: \url{https://www.nature.com/articles/s41586-025-08661-4}
\BIBentrySTDinterwordspacing

\bibitem{mipro2024}
\BIBentryALTinterwordspacing
K.~Opsahl-Ong, M.~J. Ryan, J.~Purtell, D.~Broman, C.~Potts, M.~Zaharia, and O.~Khattab, ``Optimizing instructions and demonstrations for multi-stage language model programs,'' in \emph{Proceedings of the 2024 Conference on Empirical Methods in Natural Language Processing}.\hskip 1em plus 0.5em minus 0.4em\relax Association for Computational Linguistics, 2024, pp. 9340--9366. [Online]. Available: \url{https://aclanthology.org/2024.emnlp-main.525/}
\BIBentrySTDinterwordspacing

\bibitem{dspymiprov2docs}
\BIBentryALTinterwordspacing
{DSPy}, ``{MIPROv2},'' DSPy API documentation, official living implementation documentation, accessed 2026-09-28. [Online]. Available: \url{https://dspy.ai/current/api/optimizers/MIPROv2/}
\BIBentrySTDinterwordspacing

\bibitem{tama2026}
\BIBentryALTinterwordspacing
H.~Xu, S.~Yi, T.~Lim, J.~Xu, A.~Well, C.~Mery, A.~Zhang, Y.~Zhang, H.~Ji, K.~Pingali, Y.~Leng, and Y.~Ding, ``{TAMA}: A human-{AI} collaborative thematic analysis framework using multi-agent {LLMs} for clinical interviews,'' \emph{ACM Transactions on Computing for Healthcare}, Jul. 2026. [Online]. Available: \url{https://doi.org/10.1145/3828752}
\BIBentrySTDinterwordspacing

\bibitem{sftta2025}
\BIBentryALTinterwordspacing
S.~Yi, J.~Nguyen, H.~Xu, T.~Lim, J.~Skrovan, M.~Beri, H.~Modi, A.~Well, L.~Liu, M.~Markey, and Y.~Ding, ``{SFT-TA}: Supervised fine-tuned agents in multi-agent {LLMs} for automated inductive thematic analysis,'' arXiv preprint arXiv:2509.17167, 2025. [Online]. Available: \url{https://arxiv.org/abs/2509.17167}
\BIBentrySTDinterwordspacing

\bibitem{autota2025}
\BIBentryALTinterwordspacing
S.~Yi, J.~Nguyen, H.~Xu, T.~Lim, A.~Well, M.~Markey, and Y.~Ding, ``{Auto-TA}: Towards scalable automated thematic analysis ({TA}) via multi-agent large language models with reinforcement learning,'' arXiv preprint arXiv:2506.23998v2, 2025. [Online]. Available: \url{https://arxiv.org/abs/2506.23998v2}
\BIBentrySTDinterwordspacing

\bibitem{clinicalta2026}
\BIBentryALTinterwordspacing
S.~Yi, J.~Nguyen, H.~Xu, T.~Lim, J.~Skrovan, M.~Beri, H.~Modi, A.~Well, C.~M. Mery, Y.~Zhang, M.~K. Markey, and Y.~Ding, ``Automated thematic analysis for clinical qualitative data: Iterative codebook refinement with full provenance,'' arXiv preprint arXiv:2603.08989v1, 2026. [Online]. Available: \url{https://arxiv.org/abs/2603.08989v1}
\BIBentrySTDinterwordspacing

\bibitem{positionta2025}
\BIBentryALTinterwordspacing
S.~Yi, J.~Nguyen, T.~Lim, A.~Well, J.~Skrovan, M.~Beri, Y.~Lee, K.~Radhakrishnan, L.~Liu, M.~Markey, and Y.~Ding, ``Position: Thematic analysis of unstructured clinical transcripts with large language models,'' in \emph{The Second Workshop on {GenAI} for Health: Potential, Trust, and Policy Compliance at {NeurIPS}}, 2025. [Online]. Available: \url{https://openreview.net/forum?id=XFvJSodMyi}
\BIBentrySTDinterwordspacing

\bibitem{oconnor2018imv}
\BIBentryALTinterwordspacing
R.~C. O'Connor and O.~J. Kirtley, ``The integrated motivational--volitional model of suicidal behaviour,'' \emph{Philosophical Transactions of the Royal Society B: Biological Sciences}, vol. 373, no. 1754, p. 20170268, 2018. [Online]. Available: \url{https://doi.org/10.1098/rstb.2017.0268}
\BIBentrySTDinterwordspacing

\bibitem{sloan2018wet}
\BIBentryALTinterwordspacing
D.~M. Sloan, B.~P. Marx, D.~J. Lee, and P.~A. Resick, ``A brief exposure-based treatment vs cognitive processing therapy for posttraumatic stress disorder: A randomized noninferiority clinical trial,'' \emph{JAMA Psychiatry}, vol.~75, no.~3, pp. 233--239, 2018. [Online]. Available: \url{https://doi.org/10.1001/jamapsychiatry.2017.4249}
\BIBentrySTDinterwordspacing

\bibitem{craske2008inhibitory}
\BIBentryALTinterwordspacing
M.~G. Craske, K.~Kircanski, M.~Zelikowsky, J.~Mystkowski, N.~Chowdhury, and A.~Baker, ``Optimizing inhibitory learning during exposure therapy,'' \emph{Behaviour Research and Therapy}, vol.~46, no.~1, pp. 5--27, 2008. [Online]. Available: \url{https://doi.org/10.1016/j.brat.2007.10.003}
\BIBentrySTDinterwordspacing

\bibitem{hoang2026psibench}
\BIBentryALTinterwordspacing
N.~K. Hoang, S.~Mehri, T.-A. Hsu, Y.-J. Sun, Q.~X.~N. Truong, K.~D. Doan, and D.~Hakkani-T{\"u}r, ``{PSI-Bench}: Interpretable and clinically meaningful evaluation of depression patient simulators,'' \emph{arXiv preprint arXiv:2604.25840v2}, 2026. [Online]. Available: \url{https://arxiv.org/abs/2604.25840v2}
\BIBentrySTDinterwordspacing

\bibitem{goemotionsdata}
\BIBentryALTinterwordspacing
{Google Research}, ``{GoEmotions}: Dataset documentation,'' Official Google Research repository, dataset sizes, agreement filtering, and released splits. Accessed 2026-09-28. [Online]. Available: \url{https://github.com/google-research/google-research/tree/master/goemotions}
\BIBentrySTDinterwordspacing

\bibitem{brown2020}
\BIBentryALTinterwordspacing
T.~Brown, B.~Mann, N.~Ryder, M.~Subbiah, J.~D. Kaplan, P.~Dhariwal, A.~Neelakantan, P.~Shyam, G.~Sastry, A.~Askell, S.~Agarwal, A.~Herbert-Voss, G.~Krueger, T.~Henighan, R.~Child, A.~Ramesh, D.~Ziegler, J.~Wu, C.~Winter, C.~Hesse, M.~Chen, E.~Sigler, M.~Litwin, S.~Gray, B.~Chess, J.~Clark, C.~Berner, S.~McCandlish, A.~Radford, I.~Sutskever, and D.~Amodei, ``Language models are few-shot learners,'' in \emph{Advances in Neural Information Processing Systems}, vol.~33, 2020. [Online]. Available: \url{https://proceedings.neurips.cc/paper/2020/hash/1457c0d6bfcb4967418bfb8ac142f64a-Abstract.html}
\BIBentrySTDinterwordspacing

\bibitem{goemotionslabelmap}
\BIBentryALTinterwordspacing
{Google Research}, ``{GoEmotions} label inventory: {emotions.txt},'' Official Google Research repository, ordered label file, zero-based IDs. Accessed 2026-09-28. [Online]. Available: \url{https://github.com/google-research/google-research/blob/master/goemotions/data/emotions.txt}
\BIBentrySTDinterwordspacing

\end{thebibliography}

\end{document}